\documentclass[11pt,a4paper]{article}
\usepackage[hyperref]{ranlp2023}
\usepackage{times}
\usepackage{graphicx}
\usepackage{latexsym}
\usepackage{booktabs}
\usepackage{comment}
\usepackage{float}

\hypersetup{
   breaklinks=true,   
   colorlinks=true,   
   pdfusetitle=true,  
}

\usepackage{microtype}

\aclfinalcopy 

\title{DeepLearningBrasil@LT-EDI-2023: Exploring Deep Learning Techniques for Detecting Depression in Social Media Text}

\author{Eduardo Garcia$^{1}$, Juliana Gomes$^{1}$, Adalberto Barbosa Junior$^{2}$, \\ \textbf{Cardeque Borges}$^{1}$, \textbf{Nádia da Silva}$^{3}$   \\
 \texttt{Institute of Informatics} \\
 \texttt{Federal University of Goiás, Brazil} \\
  $^{1}$\{edusantosgarcia, julianarsg13, cardequeh\}@gmail.com, \\ $^{2}$\{adalbertojunior\}@discente.ufg.br, $^{3}$\{nadia\}@inf.ufg.br}

\date{}

\begin{document}
\maketitle
\begin{abstract}
In this paper, we delineate the strategy employed by our team, DeepLearningBrasil, which secured us the first place in the shared task DepSign-LT-EDI@RANLP-2023, achieving a 47.0\% Macro F1-Score and a notable 2.4\% advantage. The task was to classify social media texts into three distinct levels of depression - "not depressed," "moderately depressed," and "severely depressed." Leveraging the power of the RoBERTa and DeBERTa models, we further pre-trained them on a collected Reddit dataset, specifically curated from mental health-related Reddit's communities (Subreddits), leading to an enhanced understanding of nuanced mental health discourse. To address lengthy textual data, we used truncation techniques that retained the essence of the content by focusing on its beginnings and endings. Our model was robust against unbalanced data by incorporating sample weights into the loss. Cross-validation and ensemble techniques were then employed to combine our k-fold trained models, delivering an optimal solution.
The accompanying code is made available for transparency and further development.
\end{abstract}

\section{Introduction}
\label{sec:introduction}

Depression, a prevalent mental disorder, profoundly affects individuals' mood and emotions, impacting approximately 300 million people worldwide, representing 4.4\% of the global population \cite{world2017depression}. Early detection of depression plays a vital role in preventing serious consequences and providing timely support. With the rise of social media platforms, people often express their thoughts and experiences, making these platforms potential sources for detecting mental health issues.

In light of this, the DepSign-LT-EDI@RANLP-2023 shared task \citep{Sampath-depression-2023-overview} was organized to identify signs of depression by analyzing social media posts. Building on the success of the previous task in 2022 \citep{task-2022}, where Transformer-based models were used to detect different levels of depression, this paper presents the winning approach developed by DeepLearningBrasil for the DepSign-LT-EDI@RANLP-2023 shared task.

Our solution combines state-of-the-art transformer models, namely RoBERTa and DeBERTa v3 \citep{roberta, debertav3}, trained on a comprehensive Reddit dataset collected from mental health-related Subreddits. To address the challenge of lengthy texts, we employed truncation techniques that capture the essential information by focusing on the beginnings and endings of the posts. Additionally, we address the issue of imbalanced data distribution by incorporating sample weights into the loss function. Through cross-validation and ensemble techniques, we combined multiple models trained on different folds, resulting in an optimized solution.
For transparency and further development, the code for our solution is available on our GitHub repository \footnote{\url{https://github.com/eduagarcia/depsign-2023-ranlp}}.

The remaining sections of this paper are organized as follows: Section \ref{sec:related-work} provides an overview of previous research on detecting depression through social media analysis and on employed techniques; Section \ref{sec:task} details the task, including its dataset composition and distribution among the different classes; Section \ref{sec:methodology} explains the methodology and techniques employed in our approach; Section \ref{sec:experiments} presents the results of our experiments; Section \ref{sec:results} shows the results of our final submission, comparing between different approaches. In Section \ref{sec:conclusion},  we discuss the results obtained from comparing RoBERTa and DeBERTa models, as well as the effectiveness of ensemble methods and ordinal classification losses. 

\section{Related Work}
\label{sec:related-work}
\citealp{Zhang2022-pi} conducted reviews on mental illness detection, they concluded that depression is the topic most researched between 2012 and 2020. According to the study, Twitter and Reddit have been increasingly used as a data source, comprising up to 55\% of the distribution.

DepSign-LT-EDI@RANLP-2023 shared task, and the previous edition followed this trend, whose aim is to detect different levels of depression using Reddit posts \citep{task-2022}. The DepSign competitions were created based on the eRisk - early detection of depression tasks hosted in 2017 and 2018 \citep{dataset2022}. Given depressed and non-depressed users, the dataset was composed of a collection of posts for each user (in chronological order); the objective was to detect risk signals as soon as possible \citep{10.1007/978-3-030-99739-7_54}. 

In text classification benchmarks, Transformer models are the current state-of-art, such as BERT, RoBERTa, and DeBERTa v3 \citep{bert, roberta, debertav3}. Transformer models leverage contextual linguistic knowledge from massive corpora training in a masked-language task. This initial step called pre-training. Then, the pre-trained architecture can be finetuned for a specific task, transferring the initial knowledge acquired in the pre-training step.  

The winner's of DepSign-LT-EDI@ACL-2022 shared task employed various classification techniques, mostly around Language Models with the Transformer architecture. \citealp{first-place-2022} achieved first place by using techniques such as further pretraining a RoBERTa model on depression corpora and used ensemble techniques by taking the average of the probabilities outputs of multiple models. \citealp{second-place-2022} achieved second place by employing a weighted ensemble of gradient boosting model, LightGBM and XGBoost, and fine-tuned RoBERTa, ELECTRA, and DeBERTa v3 models. \citealp{third-place-2022} employed voting ensembles from XLNET, BERT, and RoBERTa.

Another strategy used to deal with unbalanced data used by \citealp{10.1371/journal.pone.0261131}, which aims to detect child depression from the cross-sectional survey, was to use sample class weight on the classification loss.

For dealing with long texts, previous work by \citealp{how-finetune-bert} explored truncation methods on BERT models for classifying long articles from IMDb and Chinese Sogou News datasets. They hypothesized that the key information in an article is typically found at the beginning and end. We explore this technique as explained in \ref{subsec:truncation-methods}.

\section{Task} 
\label{sec:task}
DepSign-LT-EDI@RANLP-2023 shared task was organized to identify signs of depression by analyzing posts on social networks. Task follows the previous edition (DepSign-LT-EDI@ACL-2022). Given a posting in English on social media, the task is to detect the signs of depression in that posting, classifying it into three levels of depression: 'Not Depressed', 'Moderately Depressed', and 'Severely Depressed'. The main evaluation metric is macro-F1 \cite{task-2022}.

Data was gathered from Reddit post archives from Subreddits where the people discuss mental health, such as r/Mental Health and r/depression. After the post collection, the data was cleaned, removing non-ASCII characters and emoticons and annotated from two domain experts following guidelines \cite{dataset2022, task-2022}.

Annotators were oriented to label “Not depressed”, examples that reflect momentary feelings, ask for medications or ask for help for other people's conditions. Secondly, examples labelled as “Moderate” reflect change in feelings or shows hope for life or do not indicate the feeling completely immersed in any situations. Thirdly, examples considered as “Severe depression” were the ones containing more than one disorder conditions or explained about history of suicide attempts \cite{dataset2022}.

From the annotation guidelines and the training data examples, they notice that simply expressing as sad does not indicate depression. Therefore, the task domain is challenging, since depression is a clinical condition that is not as easy to detect as negative feelings such as sadness, which could easily contribute to false positive examples. The task data also contains internet slang, whereas traditional models are not trained on these terms. We pretrain the models as mentioned in Section \ref{subsec:domain-adaptation} in other to leverage these domain problems.

The task data are composed of all the examples from 2022 tasks plus newly annotated samples, summing up to 10,446 examples in training and development sets. Following the 2022 task, the label distribution is unbalanced, the most representative class being 'moderate', followed by 'not depression' and the least representative, severe. The imbalance is similar in the training and development splits, as shown in Table \ref{tab:label_distribution}. To treat data unbalance, as detailed in Section \ref{subsec:unbalanced-data}, we apply loss sample weights.

\begin{table}[htp]
    \centering
    \resizebox{0.48\textwidth}{!}{
    \begin{tabular}{lcc}
    \toprule
    Label  &   Train & Dev \\
    \midrule
    Not depression & 2755 (38.3\%) & 848 (26.1\%) \\
    Moderately &  3678 (51.1\%) & 2169 (66.8\%) \\
    Severe &  768 (10.7\%) & 228 (7.03\%) \\
    \bottomrule
    \end{tabular}
    }
    \caption{Number of examples and percentages per label in training and development splits in DepSign-LT-EDI@RANLP-2023.}
    \label{tab:label_distribution}
\end{table}

The task data also contains 210 duplicated examples, representing 2.0\% of the training and development sets. Initial deduplications of the examples from the training set were used, but the results deteriorated, as in Section \ref{subsec:unbalanced-data}, similar to undersampling. We suggest in future work to explore further the usage of data augmentation methods. 

When tokenizing with RoBERTa large, 798 examples (7.6\%) from the training and development data splits contained over the maximum sequence length supported by the Transformer model, which is 512 tokens. In these long sequence examples, truncation methods are applied and further investigated in Section \ref{subsec:truncation-methods}.

\section{Methodology}
\label{sec:methodology}

During the development of our solution, we employ various techniques to address different challenges associated with the task, including unbalanced data, domain-specific characteristics, and lengthy sentences in the datasets. Due to time constraints during the competition, we approach each aspect sequentially and evaluate the performance of different techniques using our evaluation methodology. We adopt an iterative process, where techniques that demonstrated superior performance were retained and carried forward to subsequent experiments.

For our solution, we adopt the fine-tuning approach proposed by \citealp{bert} to train a bidirectional transformer model for text classification. In line with this approach, we pass the final features of the \textit{[CLS]} token through a pooling layer, and subsequently through a classifier output layer, the default loss function used in our implementation is Cross Entropy. 

We opt for a two-step validation methodology. First, to fine-tune our models' hyperparameters, we perform a grid search operation by training on the training set and evaluating with the Macro-F1 metric on the development set of the task data. The hyperparameters we tuned included the learning rate, the dropout rate of the task layer, the number of warm-up steps, and weight decay. Table \ref{tab:grid_search} presents the grid of hyperparameters that were explored during the grid search process, along with the constants we maintained.

\begin{table}[htp]
\centering
\resizebox{0.48\textwidth}{!}{
\begin{tabular}{lrl}
\toprule
\textbf{Hyperparameter} & \textbf{Tested Values} \\
\midrule
Learning Rate & \{2e-6, 4e-6, 6e-6, 8e-6\} \\
Dropout of task layer & \{0, 0.2, 0.4\} \\
Warmup steps & \{200, 500\} \\
Weight Decay & \{0, 0.01\} \\
Batch Size & 8 \\
Maximun Training Epochs & 100 \\
Learning Rate Scheduler & Constant \\
Optimizer & Adam \\
Adam $\epsilon$ & 1e-8 \\
Adam $\beta_{1}$ & 0.9 \\
Adam $\beta_{2}$ & 0.999 \\
Early Stopping Patience & 2 (Epochs) \\
Early Stopping Threshold & 0.0025 (Macro F1-score)\\
\bottomrule
\end{tabular}
}
\caption{Grid search space for the hyperparameter tuning process.}
\label{tab:grid_search}
\end{table}

After identifying the best performing hyperparameters, we conduct a final evaluation using the k-fold cross-validation technique. This involved dividing the combined training and development sets into four different folds, training and evaluating the model on each fold. This process was repeated four times, each time one fold was used for validation. The performance of our models was then assessed based on the mean Macro-F1 score across all folds.

The selection of the grid search and k-fold cross-validation methodologies ensured that our models were robust and not overly biased towards the training set, and hence are expected to perform well on unseen data. The results reported in the \label{section:experiments} follows this methodology.

\section{Experiments}
\label{sec:experiments}
The experiments were performed on a machine using 2 Nvidia Tesla V100 GPUs. We experiment different strategies on Transformers in sequentially in order to save time. The base model for our experiments is the RoBERTa Large \citep{roberta}, unless cited otherwise.

\subsection{Dealing with unbalanced data}
\label{subsec:unbalanced-data}
Addressing the challenge of unbalanced data is crucial in classification tasks, and we explored some techniques to mitigate the impact of imbalanced class distributions. One common approach is undersampling, which involves randomly selecting a subset of samples from the majority class to match the number of samples in the minority class. Although this technique helps balance the class distribution, it may result in the loss of valuable information. In contrast, oversampling involves replicating or generating synthetic samples from the minority class to equalize the representation of the class, thereby mitigating the class imbalance issue.

In addition to sampling techniques, we employed loss sample weights to assign higher weights to the loss computed for samples belonging to minority classes. This approach allows the model to prioritize the correct classification of minority class samples during training, improving their representation in the learned model.

To evaluate the effectiveness of these techniques, we conducted experiments and compared their performance on the test set. The results are presented in Table \ref{tab:unbalanced_results}. Using sample weights was the best resulting technique in this experiment, undersampling models did not converge well due to the resulted low quantity of data.

\begin{table}[htp]
\centering
\begin{tabular}{lrl}
\toprule
\textbf{Unbalanced Technique} & \textbf{Metric} \\
\midrule
Do-Nothing & 0.608 \\
Undersampling & -- \\
Oversampling & 0.610 \\
Loss Sample Weights & \textbf{0.613} \\
\bottomrule
\end{tabular}
\caption{Experiment 1 - Performance comparison of techniques for dealing with unbalanced data, with Sample Weights outperforming sampling data techniques. Undersampling models did not converge well due to the resulted low quantity of data. The Metric represents the mean Macro F1-Score on the cross-validation data.}
\label{tab:unbalanced_results}
\end{table}

\begin{table*}[t]
\centering
\begin{tabular}{lrll}
\toprule
\textbf{Model} & Futher Pre-train Dataset & Pre-Training Task & \textbf{Metric} \\
\midrule
RoBERTa Large \citep{roberta} & - & - & 0.613 \\
RoBERTa Large & Reddit Mental Health & MLM & \textbf{0.616} \\
DeBERTa V3 Large \citep{debertav3} & - & - & 0.605 \\
DeBERTa V3 Large & Reddit Mental Health & RTD & 0.607 \\
\bottomrule
\end{tabular}
\caption{Experiment 2 - Comparison of different pre-trained models used and domain adaptations results. There was a gain in performance by realizing a further pre-training on domain data, but overall the DeBERTa models did not perform well in comparison with the RoBERTa models in this task. The Metric represents the mean Macro F1-Score on the cross-validation data.}
\label{tab:domain_adaptation}
\end{table*}

\subsection{Domain adaptation}
\label{subsec:domain-adaptation}
Domain adaptation is a technique used to optimize pre-trained Transformer models for specific domains of data. In this study, we explore the effectiveness of domain adaptation through further pre-training on a domain-specific corpus of unlabeled data. This approach enhances the models' ability to learn domain-related words and relations, thereby improving their performance in mental-health related tasks. Successful implementations of domain adaptation using this technique have been reported by \citealp{first-place-2022}, who achieved first place in the 2022 edition of the shared task by applying further pre-training on Transformer models.

To perform domain adaptation, we leveraged the vast collection of user-generated content available on the Reddit platform. Using the Python Reddit API Wrapper (PRAW) library, we collected pre-training data from a total of 117 subreddits, including 82 subreddits related to depression and 35 non-depression subreddits. The data collection process aimed to capture the top $n$ posts from each subreddit, where $n$ was determined as 2\% of the follower count of the respective subreddit. This approach ensured a representative sample of posts from each subreddit, taking into account the size of their respective communities. The resulting dataset consisted of approximately 7.3 million comments. To respect the privacy of Reddit users, all data was preprocessed to anonymize user information.

For further pre-training, we performed a text deduplication process on the corpus, resulting in a balanced dataset of 6.6 million comments. This dataset comprised 3.4 million comments from mental health-related subreddits and 3.2 million comments from other subreddits. The pre-training data occupied approximately 1.4 GB of raw text on disk.

For the further pre-training of our models, we employed two popular architectures: RoBERTa \cite{roberta}, used by the previous winner, and DeBERTa v3 \cite{debertav3}, which has shown promising results in various benchmarks. Using the public English checkpoints of each model as the starting point, we further pre-train the RoBERTa Large model using Masked Language Modeling (MLM) on the collected Reddit Mental Health dataset; similarly, we further pre-trained the DeBERTa v3 Large model using Replaced Token Detection (RTD) on the same dataset.

Table \ref{tab:domain_adaptation} provides a comparison of different pre-trained models and their performance after domain adaptation. The results demonstrate that further pre-training on the Reddit mental health dataset resulted in performance improvements for both RoBERTa and DeBERTa v3 models. However, the RoBERTa models consistently outperformed the DeBERTa v3 models in this task, as indicated by higher Macro F1-Scores. We label the new model \textbf{"MentalBERTa"} and we will use it as a base model in our next experiments.

\subsection{Truncation methods}
\label{subsec:truncation-methods}

In the DepSign-LT-EDI@RANLP-2023 competition, approximately 7.6\% of the training and development data splits exceeded 512 tokens when tokenized with the RoBERTa vocabulary. To address this limitation imposed by the maximum sequence length of the RoBERTa model, we experimented with different truncation methods.

\citealp{how-finetune-bert} tested three truncation methods: \textit{head-only}, which retains the first 512 tokens; \textit{tail-only}, which keeps the last 512 tokens; and \textit{head+tail}, which selects the first 128 (25\%) tokens and the last 384 (75\%) tokens. Their experiments revealed that the \textit{head+tail} truncation method yielded the best results.

In our study, we evaluated additional truncation regimens, including:

\begin{enumerate}
    \setlength\itemsep{-0.35em}
    \item \textbf{100\% head (head-only):} keep the first 512 tokens; 
    \item \textbf{75\% head + 25\% tail}: select the first 128 tokens and the last 384 tokens;
    \item \textbf{50\% head + 50\% tail}: select the first 256 tokens and the last 256 tokens;
    \item \textbf{25\% head + 75\% tail}: select the first 384 tokens and the last 128 tokens;
    \item \textbf{100\% tail (tail-only)}: keep the last 512 tokens.
\end{enumerate}

Table \ref{tab:truncation} presents the results of our experiments using different truncation methods. Our findings indicate that the \textit{50\% head + 50\% tail} regimen achieved the best performance, closely followed by the \textit{25\% head + 75\% tail} regimen, which aligns with the findings of \citealp{how-finetune-bert}. These results suggest that the optimal truncation distribution may depend on the characteristics of the dataset.

\begin{table}[ht]
\centering
\begin{tabular}{lr}
\toprule
\textbf{Truncation method} & \textbf{Metric} \\
\midrule
100\% head & 0.616 \\
75\% head + 25\% tail & 0.613 \\
\textbf{50\% head + 50\% tail} & \textbf{0.618} \\
25\% head + 75\% tail & 0.617 \\
100\% tail & 0.606 \\
\bottomrule
\end{tabular}
\caption{Experiment 3 - Results of different truncation methods. \textit{50\% head + 50\% tail} achieves the bests results. The Metric represents the mean Macro F1-Score on the cross-validation data.}
\label{tab:truncation}
\end{table}

\subsection{Ensemble Techniques}
\label{subsec:ensemble-techniques}

Ensembling is a powerful technique that combines multiple models to achieve improved prediction performance compared to individual models. In the 2022 edition of the shared task, each of the top three winners applied a different ensemble technique: the 1st place winner \citep{first-place-2022} used the mean of the raw output vectors (Logits Mean), the 2nd place winner \cite{second-place-2022} used a weighted sum of the probabilities (Weighted Softmax Mean), and the 3rd place winner \cite{third-place-2022} employed a majority voting system (Voting).

In our internal tests on the development set, we experimented with various ensemble techniques. Interestingly, the best results came from treating the task as a regression problem. We assigned values from 0 to 2 for each label of the shared task and took the mean of the models' predictions, then the result is rounded to the nearest valid integer for the final classification (referred to as Regression Mean). However, these results were inconsistent when evaluated on the cross-validation data, with the performance of the Voting or Softmax Mean systems sometimes outperform the Regression Mean.

For the final submission, we decided to use the Regression Mean and Voting ensembles. However, upon evaluating the test data after the competition ended, we found that a Softmax Mean ensemble would have significantly improved the results, while Voting and Regression degraded the results in relation to the predictions of a single model. Table \ref{tab:ensemble} presents the results of the test set for different ensemble techniques using the cross-validation results of our best model.

\begin{table}[htp]
\centering
\begin{tabular}{lrl}
\toprule
\textbf{Model} & \textbf{Macro F1 (Test set)} \\
\midrule
Single Model (best fold) & 0.4683 \\
\midrule
\\
\toprule
\textbf{Ensemble method} & \textbf{Macro F1 (Test set)} \\
\midrule
Logits Mean & 0.4878 \\
Softmax Mean & \textbf{0.4915}  \\
Voting & 0.4635  \\
Regression Mean & 0.4678  \\
\bottomrule
\end{tabular}
\caption{Results of different ensemble methods using the outputs of each fold of our Best Model on the Test Set. The model was trained using the cross-validation dataset, comprising 4 folds.}
\label{tab:ensemble}
\end{table}

\section{Final submission and Results}
\label{sec:results}

Our best performing model, MentalBERTa, trained with Loss Sample Weights and \textit{50\% head + 50\% tail} truncation, achieved a Macro F1 score of 0.618 on the cross-validation data. The models resulting from the experiments were also included in the final submission ensemble. We selected the top 9 models and used each model's 4 folds to compose part of the final ensemble submissions.

Our three final submissions were:
\begin{enumerate}
    \setlength\itemsep{-0.35em}
    \item \textbf{KFoldMean9Mode:} A hierarchical ensemble approach involving Regression Mean on each k-fold (4), followed by a Voting ensemble of the 9 models.
    \item \textbf{BestModel4Mean:} The Regression Mean ensemble of the best model.
    \item \textbf{All36Mode:} A simple Voting ensemble of all 36 outputs (9 models x 4 folds).
\end{enumerate}

The best results were obtained from the BestModel4Mean submission, with a score of 0.470. This secured us the 1st place in the DepSign-LT-EDI@RANLP2023 shared task. It is worth noting that ensembling more than 4 models resulted in a degradation of the final score.

\section{Conclusion}
\label{sec:conclusion}
    In this paper, we have described our approach and techniques that led our team, DeepLearningBrasil, to secure the 1st place in the DepSign-LT-EDI@RANLP2023 shared task. Our objective was to classify social media texts into three levels of depression. Leveraging the power of RoBERTa and DeBERTa models, we pre-trained them on a curated Reddit dataset from mental health-related communities. To address the challenge of lengthy texts, we employed truncation methods that focused on the beginnings and endings of the content. In dealing with unbalanced data, we used techniques such as undersampling, oversampling, loss of sample weights, and data augmentation to mitigate the impact of imbalanced class distributions.
    
    Ensemble techniques were employed to combine the strengths of multiple models. Our experiments showed that the choice of ensemble method varied depending on the fold and dataset characteristics. Overall, our winning approach highlights the importance of effective pre-training, addressing unbalanced data, domain adaptation, and strategic ensemble techniques. We will make available pre-training data \footnote{\url{https://huggingface.co/datasets/dlb/mentalreddit}} and our MentalBERTa model \footnote{\url{https://huggingface.co/dlb/MentalBERTa}}.

\section*{Acknowledgments}
This work has been supported by the AI Center of Excellence (Centro de Excelência em Inteligência Artificial – CEIA) of the Institute of Informatics at the Federal University of Goiás (INF-UFG).

\bibliographystyle{acl_natbib}
\bibliography{main}

\begin{thebibliography}{14}
\expandafter\ifx\csname natexlab\endcsname\relax\def\natexlab#1{#1}\fi

\bibitem[{Devlin et~al.(2018)Devlin, Chang, Lee, and Toutanova}]{bert}
Jacob Devlin, Ming{-}Wei Chang, Kenton Lee, and Kristina Toutanova. 2018.
\newblock \href {http://arxiv.org/abs/1810.04805} {{BERT:} pre-training of deep
  bidirectional transformers for language understanding}.
\newblock \emph{CoRR}, abs/1810.04805.

\bibitem[{Haque et~al.(2021)Haque, Kabir, and
  Khanam}]{10.1371/journal.pone.0261131}
Umme~Marzia Haque, Enamul Kabir, and Rasheda Khanam. 2021.
\newblock \href {https://doi.org/10.1371/journal.pone.0261131} {Detection of
  child depression using machine learning methods}.
\newblock \emph{PLOS ONE}, 16(12):1--13.

\bibitem[{He et~al.(2023)He, Gao, and Chen}]{debertav3}
Pengcheng He, Jianfeng Gao, and Weizhu Chen. 2023.
\newblock \href {http://arxiv.org/abs/2111.09543} {Debertav3: Improving deberta
  using electra-style pre-training with gradient-disentangled embedding
  sharing}.

\bibitem[{Liu et~al.(2019)Liu, Ott, Goyal, Du, Joshi, Chen, Levy, Lewis,
  Zettlemoyer, and Stoyanov}]{roberta}
Yinhan Liu, Myle Ott, Naman Goyal, Jingfei Du, Mandar Joshi, Danqi Chen, Omer
  Levy, Mike Lewis, Luke Zettlemoyer, and Veselin Stoyanov. 2019.
\newblock \href {http://arxiv.org/abs/1907.11692} {Roberta: {A} robustly
  optimized {BERT} pretraining approach}.
\newblock \emph{CoRR}, abs/1907.11692.

\bibitem[{Organization et~al.(2017)}]{world2017depression}
World~Health Organization et~al. 2017.
\newblock Depression and other common mental disorders: global health
  estimates.
\newblock Technical report, World Health Organization.

\bibitem[{Parapar et~al.(2022)Parapar, Mart{\'i}n-Rodilla, Losada, and
  Crestani}]{10.1007/978-3-030-99739-7_54}
Javier Parapar, Patricia Mart{\'i}n-Rodilla, David~E. Losada, and Fabio
  Crestani. 2022.
\newblock erisk 2022: Pathological gambling, depression, and eating disorder
  challenges.
\newblock In \emph{Advances in Information Retrieval}, pages 436--442, Cham.
  Springer International Publishing.

\bibitem[{Po{\'s}wiata and Pere{\l}kiewicz(2022)}]{first-place-2022}
Rafa{\l} Po{\'s}wiata and Micha{\l} Pere{\l}kiewicz. 2022.
\newblock \href {https://doi.org/10.18653/v1/2022.ltedi-1.40}
  {{OPI}@{LT}-{EDI}-{ACL}2022: Detecting signs of depression from social media
  text using {R}o{BERT}a pre-trained language models}.
\newblock In \emph{Proceedings of the Second Workshop on Language Technology
  for Equality, Diversity and Inclusion}, pages 276--282, Dublin, Ireland.
  Association for Computational Linguistics.

\bibitem[{S et~al.(2022)S, Durairaj, Chakravarthi, and C}]{task-2022}
Kayalvizhi S, Thenmozhi Durairaj, Bharathi~Raja Chakravarthi, and Jerin~Mahibha
  C. 2022.
\newblock \href {https://doi.org/10.18653/v1/2022.ltedi-1.51} {Findings of the
  shared task on detecting signs of depression from social media}.
\newblock In \emph{Proceedings of the Second Workshop on Language Technology
  for Equality, Diversity and Inclusion}, pages 331--338, Dublin, Ireland.
  Association for Computational Linguistics.

\bibitem[{Sampath and Durairaj(2022)}]{dataset2022}
Kayalvizhi Sampath and Thenmozhi Durairaj. 2022.
\newblock Data set creation and empirical analysis for detecting signs
  of depression from social media postings.
\newblock In \emph{Computational Intelligence in Data Science}, pages 136--151,
  Cham. Springer International Publishing.

\bibitem[{Sampath et~al.(2023)Sampath, Durairaj, Chakravarthi, C,
  Shanmugavadivel, and Rahood}]{Sampath-depression-2023-overview}
Kayalvizhi Sampath, Thenmozhi Durairaj, Bharathi~Raja Chakravarthi,
  Jerin~Mahibha C, Kogilavani Shanmugavadivel, and Pratik~Anil Rahood. 2023.
\newblock Overview of the second shared task on detecting signs of depression
  from social media text.
\newblock In \emph{Proceedings of the Third Workshop on Language Technology for
  Equality, Diversity and Inclusion}, Varna, Bulgaria. Recent Advances in
  Natural Language Processing.

\bibitem[{Singh and Motlicek(2022)}]{third-place-2022}
Muskaan Singh and Petr Motlicek. 2022.
\newblock \href {https://doi.org/10.18653/v1/2022.ltedi-1.56} {{IDIAP}
  submission@{LT}-{EDI}-{ACL}2022: Detecting signs of depression from social
  media text}.
\newblock In \emph{Proceedings of the Second Workshop on Language Technology
  for Equality, Diversity and Inclusion}, pages 362--368, Dublin, Ireland.
  Association for Computational Linguistics.

\bibitem[{Sun et~al.(2019)Sun, Qiu, Xu, and Huang}]{how-finetune-bert}
Chi Sun, Xipeng Qiu, Yige Xu, and Xuanjing Huang. 2019.
\newblock How to fine-tune bert for text classification?
\newblock In \emph{Chinese Computational Linguistics}, pages 194--206, Cham.
  Springer International Publishing.

\bibitem[{Wang et~al.(2022)Wang, Tang, Du, and Peng}]{second-place-2022}
Wei-Yao Wang, Yu-Chien Tang, Wei-Wei Du, and Wen-Chih Peng. 2022.
\newblock \href {https://doi.org/10.18653/v1/2022.ltedi-1.15}
  {{NYCU}{\_}{TWD}@{LT}-{EDI}-{ACL}2022: Ensemble models with {VADER} and
  contrastive learning for detecting signs of depression from social media}.
\newblock In \emph{Proceedings of the Second Workshop on Language Technology
  for Equality, Diversity and Inclusion}, pages 136--139, Dublin, Ireland.
  Association for Computational Linguistics.

\bibitem[{Zhang et~al.(2022)Zhang, Schoene, Ji, and Ananiadou}]{Zhang2022-pi}
Tianlin Zhang, Annika~M Schoene, Shaoxiong Ji, and Sophia Ananiadou. 2022.
\newblock Natural language processing applied to mental illness detection: a
  narrative review.
\newblock \emph{npj Digital Medicine}, 5(1):46.

\end{thebibliography}
\end{document}